\documentclass[10pt,twocolumn,letterpaper]{article}

\usepackage[pagenumbers]{cvpr} %

\usepackage{graphicx}
\usepackage{amsmath}
\usepackage{amssymb}
\usepackage{booktabs}
\usepackage[usenames,dvipsnames]{xcolor}
\usepackage{hhline}
\usepackage{xspace}

\usepackage[pagebackref,breaklinks,colorlinks]{hyperref}

\usepackage[capitalize]{cleveref}
\crefname{section}{Sec.}{Secs.}
\Crefname{section}{Section}{Sections}
\Crefname{table}{Table}{Tables}
\crefname{table}{Tab.}{Tabs.}

\def\cvprPaperID{84} %
\def\confName{CVPR}
\def\confYear{2022}

\DeclareMathOperator*{\argmin}{arg\,min}

\newcommand{\topic}[1]{\vspace{1mm}\noindent\textbf{#1}}

\newcommand{\dataset}{COSy\xspace}
\newcommand{\datasetlong}{Common Objects Synthetic\xspace}

\begin{document}

\title{De-rendering 3D Objects in the Wild}

\author{Felix Wimbauer$^{1, 2}$ \hspace{1cm} Shangzhe Wu$^{2}$ \hspace{1cm} Christian Rupprecht$^{2}$\\
$^1$ Technical University Munich, $^2$ University of Oxford\\
{\tt\small wimbauer@in.tum.de \{szwu, chrisr\}@robots.ox.ac.uk}
}
\twocolumn[\maketitle\vspace{-3em}
\begin{center}
    \includegraphics[width=0.95\textwidth]{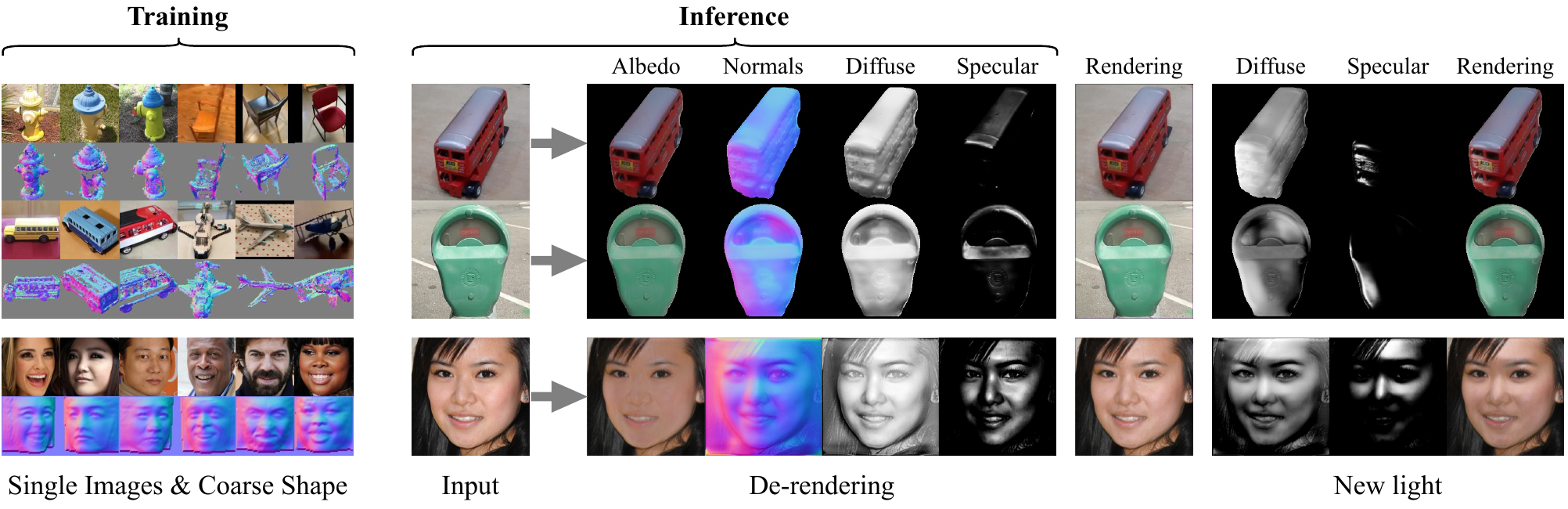}
\end{center}\vspace{-1.5em}

\captionof{figure}{\textbf{De-rendering 3D objects from in-the-wild images.} Left: Our method is trained on unconstrained  images with coarse shape estimates which can be obtained using off-the-shelf methods or classical structure from motion. Right: Our method de-renders an image into precise shape, material (diffuse albedo, specularity, specular intensity), and lighting (direction, intensity). The image can then be re-rendered and re-lit based on our illumination model. The model generalizes to objects beyond the training categories (\eg parkingmeter).}
\label{fig:teaser}
\bigbreak]
\begin{abstract}
With increasing focus on augmented and virtual reality (XR) applications comes the demand for algorithms that can lift objects from images into representations that are suitable for a wide variety of related 3D tasks.
Large-scale deployment of XR devices and applications means that we cannot solely rely on supervised learning, as collecting and annotating data for the unlimited variety of objects in the real world is infeasible.
We present a weakly supervised method that is able to decompose a single image of an object into shape (depth and normals), material (albedo, reflectivity and shininess) and global lighting parameters. 
For training, the method only relies on a rough initial shape estimate of the training objects to bootstrap the learning process. 
This shape supervision can come for example from a pretrained depth network or---more generically---from a traditional structure-from-motion pipeline.
In our experiments, we show that the method can successfully de-render 2D images into a decomposed 3D representation and generalizes to unseen object categories.
Since in-the-wild evaluation is difficult due to the lack of ground truth data, we also introduce a photo-realistic synthetic test set that allows for quantitative evaluation.
Please find our project page at: \footnotesize{\url{https://github.com/Brummi/derender3d}}\vspace{-\baselineskip}
\end{abstract}
\section{Introduction}

From a single 2D image, humans can easily reason about the underlying 3D properties of an object, such as the 3D shape, the surface material and its illumination properties.
Being able to infer ``object intrinsics'' from a single image has been a long standing goal in Computer Vision and is often referred to as ``inverse rendering'' or ``\textit{de-rendering}'' as it reverses the well-known rendering step of Computer Graphics, where an image is generated from a similar set of object and material descriptors.

De-rendering an image into its physical components, not only plays an important role for general image understanding, but is also key to many applications, such as Augmented/Virtual Reality (XR) and Visual Effects (VFX).
In these applications, a decomposed 3D representation can be used to increase the realism by enabling post-processing steps, such as relighting or changing the texture or material properties, which further blurs the line between real and synthetic objects in these environments.

As XR is moving from research and commercial use to consumer devices, a de-rendering method should work on a wide variety of images in the wild to allow a broad adoption of these technologies.
While the history of image decomposition literature is long~\cite{horn1975obtaining, horn89shape}, recent learning-based approaches have demonstrated this capability on specific categories, such as human portraits~\cite{TotalRelighting} and synthetic ShapeNet objects~\cite{ShapenetIntrinsics}, by training on ground-truth data, often obtained using synthetic models or sophisticated light stage capturing systems.
However, obtaining large-scale ground-truth material and illumination annotations for general objects ``in the wild'' is much more challenging and infeasible to collect for \textit{all} objects.
Models trained on synthetic data often lack sufficient realism, resulting in poor transfer to real images.
Models trained on real data usually focus on a single category (\eg faces or birds \cite{Wu_2020_CVPR,kanazawa18learning,goel20shape,li2020online,li20self-supervised}) and do not generalize to new classes.

On the other hand, another line of research that has recently gained interest, aims to learn 3D objects in an unsupervised or weakly-supervised fashion, without relying on explicit 3D ground-truth~\cite{Wu_2020_CVPR, kulkarni2019canonical, kanazawa18learning, li20self-supervised, moniz2018unsupervised}.
Although impressive results have been demonstrated in reconstructing 3D shapes of simple objects, few of the methods have considered also recovering specular surface materials as this introduces even more ambiguities to the model.
Furthermore, they are generally restricted to a single category.

In this paper, we explore the problem of learning non-Lambertian intrinsic decomposition from in-the-wild images without relying on explicit ground-truth annotations.
In particular, we introduce a method that capitalizes on the coarse 3D shape reconstructions obtained from unsupervised methods and learns to predict a refined shape as well as further decomposes the material into albedo and specular components, given a collection of single-view images.

At the core of the method lies an image formation process that renders the image from its individual components. The model then learns to decompose the image through a reconstruction objective. 
Since this formulation is highly ambiguous, the model relies on several additional cues to enable learning a meaningful decomposition.
We bootstrap the training using a coarse estimate of the initial shape. This estimate can come from a variety of sources. For datasets such as Co3D~\cite{reizenstein2021common}, where multi-view information is available, we rely on traditional structure-from-motion pipelines (\eg COLMAP~\cite{schoenberger2016sfm}). For specific categories such as faces, existing specialized unsupervised methods can be used to obtain a coarse initial shape estimate.
We present a simple method that estimates initial material and light properties using the coarse shape, the input image and a simple lighting model.
We can then facilitate learning by using the coarse estimates as initial supervisory signals, which avoids many degenerate solutions that would fulfill the reconstruction objective alone.
Finally, to further improve the quality of the decomposition, we introduce a third objective, where the image is rendered with randomized light parameters, and a discriminator helps to ensure realistic reconstructions.

While we do need (pseudo) supervision of the coarse shape during training, the final model can directly decompose an input image without any other input. 
We show that our model produces accurate and convincing image decompositions that improve the state of the art and even generalizes beyond the categories of objects it was trained on.
In our experiments, we show that the model works on a wide variety of objects from different datasets. 
However, as this is the first method to tackle de-rendering in the wild, there is currently no suitable benchmark to quantitatively evaluate the quality of the decomposition. 
We thus also introduce a synthetic benchmark dataset, using photo-realistic rendering of 10 objects from several viewpoints. 
Each image is associated with ground truth per-pixel material properties and lighting information that allows us to directly evaluate the decomposition.
The new dataset, code and trained models will be published together with the paper.

\section{Related Work}
This work studies the problem of learning to de-render images of general objects ``in the wild'', which lies in the intersection of several fields of Computer Vision and Computer Graphics.
In this section, we will first discuss the relevant work on Intrinsic Image Decomposition and Inverse Rendering from multiple images, and direct supervision as well as recent unsupervised approaches.

\topic{Instrinsic Image Decomposition.}
Intrinsic Image Decomposition is a classic task, where the main goal is to factorize an image into a reflectance image and a shading image, \ie separating the true surface color from lighting effects.
Since this is a highly ill-posed task, traditional methods often rely on additional heuristics and priors.
The classic Retinex algorithm~\cite{Land71} assumes that small variations in image intensity result from shading whereas abrupt changes reveal the true reflectance.
Many other priors have also been explored over the past few decades, such as global sparsity constraints on the reflectance~\cite{shen08intrinsic, Rother11Recovering, Shen11Intrinsic, Garces12}, and explicit geometric constraints on shading assuming Lambertian surface~\cite{BarronTPAMI2015, laffont2012coherent}.
Recently, researchers have also studied learning-based approaches, by training on synthetic data~\cite{janner2017intrinsic, Liu2020Unsupervised} or multi-illumination images~\cite{Li18Learning, liu2020factorize}.
In this work, we borrow ideas from this area to constrain the albedo extraction, but aim at decomposing the image into explicit material, shape and lighting factors rather than a single shading map, as this allows for relighting and re-rendering.

\topic{Supervised Inverse Rendering.} %
Next, we will focus on inverse rendering methods that recover shape, material and illumination from images.
Classical Shape-from-Shading approaches assume Lambertian surface properties~\cite{horn1975obtaining, horn89shape}.
Photometric Stereo techniques~\cite{goldman2009shape,alldrin2008photometric}
recover shape, BRDF material and lighting by solving an optimization problem, given multiple images of a scene captured under various lighting conditions and/or from multiple viewpoints.
This has been extended with learning-based approaches~\cite{yu19inverserendernet, YuSelfRelight20, bi2020deep, Bi2020DeepECCV, Boss2020brdf, zhang2021physg} and recently with implicit (neural) representations~\cite{yariv2020multiview, boss2021nerd, nerv2021, nerfactor}.
While most of these methods still require multiple images at inference, some have learned priors from multiple views that can be used for single image inference~\cite{kulkarni2015deep, ma2018single}.
However, capturing multiple images with controlled lighting for either training or inference is challenging and difficult to apply to objects ``in the wild'', which is main target of this paper.

The de-rendering task can also be learned with direct supervision, often using synthetic data, like ShapeNet~\cite{shapenet2015} objects~\cite{ShapenetIntrinsics, liu2017material, Boss2020brdf, chen2019learning, chen2021dib}, synthetic faces/bodies~\cite{sfsnetSengupta18, hou2021towards, lagunas2021single}, near-planar surfaces~\cite{li2018materials}, indoor scenes~\cite{li2020inverse} or other synthetic objects~\cite{janner2017intrinsic, li2018learning, sang2020single}.
However, generating large-scale realistic synthetic data that captures the level of complexity of the real world is challenging, and hence it remains questionable how well these methods generalize to real images.
As inverse rendering and relighting of faces and persons is particularly useful, relighting datasets for real faces have been collected using light stage setups~\cite{TotalRelighting, nestmeyer2020learning, sun2019single, tajima2021relighting}.
This approach, however, is not feasible for general objects.

\topic{Unsupervised Inverse Rendering.}
Recently, there has been an increasing interest in developing unsupervised or weakly supervised methods for inverse rendering tasks.
Several works have attempted to learn 3D shapes of object categories, such as faces and birds, from only single-view image collections~\cite{kanazawa18learning, Wu_2020_CVPR, li20self-supervised, goel20shape, ye21shelf-supervised}, with weak supervision such as 2D keypoints, masks, category template shapes or assumptions like symmetry.
Most of these focus on shape learning and do not tackle material and lighting decomposition specifically or  assume a simple Lambertian shading model.
Wu~\etal~\cite{wu2021derender} recovers shape, shiny material and environment lighting, but focuses only on a single specific type of object---vases---and assumes rotational symmetry.

Unlike all these approaches, this work aims at recovering specular material and illumination on \emph{general objects} from images in the wild, with only coarse geometry estimations during training, that can be obtained from existing methods.

\section{Method}
\begin{figure*}
  \centering
  \includegraphics[width=\linewidth]{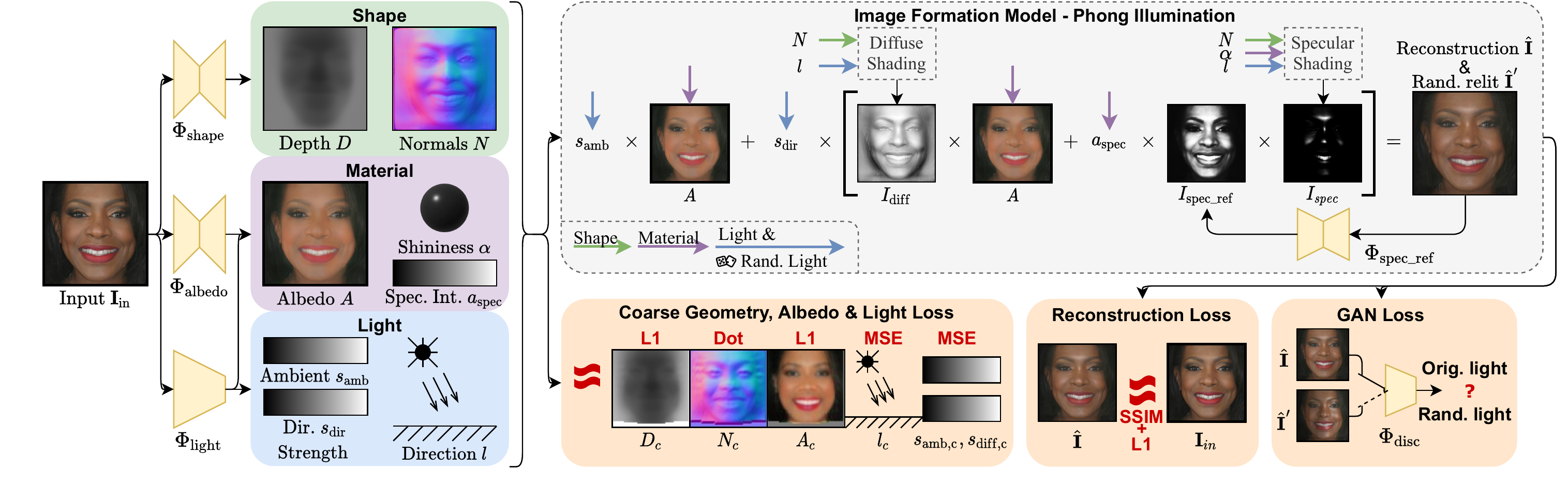}
  \caption{\textbf{Model architecture.} Several networks decompose the input image $\mathbf{I}_\mathrm{in}$ into shape (depth $D$ and normals $N$), material (albedo $A$, shininess $\alpha$ and specular intensity $a_\mathrm{spec}$), and light (ambient and directional strength $s_\mathrm{amb}, s_\mathrm{dir}$ and light direction $l$). %
  To allow for complex lighting effects, we also introduce a specularity refinement step.
  Training combines three different loss terms: 1.\ a loss on the decomposition using coarse estimates 2.\ a reconstruction loss, and 3.\ a discriminator loss for a randomly relit image.}
   \vspace{-.25cm}
  \label{fig:method}
\end{figure*}

In this section we will describe the model and training scheme of our method. \cref{fig:method} shows an overview of the decomposition, training procedure and losses.

\subsection{Rendering - Image Formation Model}

Essential to our method is modeling the image formation process, \ie rendering an image from its intrinsic components.
Our method learns to invert this process---\textit{de-rendering} the image---by extracting the intrinsic components from an input image $\textbf{I}_\mathrm{in} \in [0, 1]^{3 \times H \times W}$.%

While the rendering process is usually deterministic, it is however surjective.
This means that due to the highly complex nature of the image formation process, the inverse is ambiguous and many different combinations of intrinsic materials map to the same image. 

We deal with the highly ambiguous inverse rendering step in three ways.
First, we make reasonable assumptions about the object's material that drastically simplify the rendering process. Second, we leave enough flexibility in the rendering process, such that the model can \textit{learn} to overcome the approximations used in the first step. Finally, we use coarse shape supervision using traditional methods or existing, object-specific solutions to bootstrap the learning process and to avoid degenerate solutions. 

\topic{Shape.}
As the shape of an object has a strong influence on its shading, we will directly link these two components.
To compute per-pixel shading, we require a per-pixel normal map $N \in [-1, 1]^{3 \times H \times W}$.
Given an image, directly predicting a normal map with a neural network is problematic, as there is no incentive to adhere to a global shape.
Thus, we compute the normal map $N_D$ from a depth map $D \in [d_{\mathrm{min}}, d_\mathrm{max}]^{H \times W}$. 
However, fine geometric details (\eg scratches or small reliefs) have a strong influence on the normals, but little on the global shape.
Thus, we predict both a depth map $D$ (and compute the corresponding normal map $N_\mathrm{D}$) and a refinement normal map $N_{\mathrm{ref}}$ from the image, and combine them:
\begin{equation}\label{eq:normal_ref}
    N = \frac{N_\mathrm{D} + N_\mathrm{ref}}{\lVert N_\mathrm{D} + N_\mathrm{ref} \rVert}
\end{equation}

\topic{Light \& Material.}
Using a very expressive lighting model would allow us to capture highly complex effects for photo-realistic rendering.
However, we find that such models add significant difficulty to the learning of inverse tasks without further supervision.
To model the lighting, we hence rely on Phong Illumination~\cite{phong1975illumination}, which considers ambient, diffuse, and specular light components.
Additionally, we make the following assumptions.
We can observe during training that the shading from the one dominant light source (\eg the sun) is a very important hint for the model.
Further, multiple light sources would introduce more ambiguity, potentially harming the correctness of the predicted geometry.
Therefore, we model the light as a single directional light source and a global ambient light, both emitting perfectly white light.
It is parameterized by ambient and directional strength $s_{\mathrm{amb}}, s_{\mathrm{dir}} \in [0, 1]$, and a light direction $l \in SO(3)$.
For both terms, we use a combined per-pixel albedo map $A \in [0, 1]^{3\times H\times W}$.
Specularity is a very complex lighting effect and therefore difficult to extract from a single image.
To keep the complexity tractable, we use a global shininess value $\alpha \in [0, \alpha_{\mathrm{max}}]$ and a global specularity intensity $a_\mathrm{spec} \in [0, 1]$
for the whole object.
Summarizing, we represent the lighting as $L = (s_\mathrm{amb}, s_\mathrm{diff}, l)$ and the intrinsic material properties as $(A, \alpha, a_\mathrm{spec})$.

We obtain an image $\hat{\textbf{I}}$ from shape, material, and lighting through the following rendering equation, where $u \in \Omega = \{1, \ldots, H\}\times\{1, \ldots, W\}$ represents a pixel location.
\begin{equation}
    \hat{\textbf{I}}_u = \tau (\underbrace{s_\mathrm{amb} A_u}_{\text{ambient term}} + 
    s_\mathrm{dir} (\underbrace{N_u^T l A_u}_{\text{diffuse term}} + 
    \underbrace{a_\mathrm{spec}\left(r^T v\right)^\alpha}_{\text{specular term}})),
\end{equation}
where $r$ is the reflected light direction, and gamma function $\tau(\mathbf{I}_u)=\mathbf{I}_u^{1 / \gamma},\; \gamma=2.2$ approximates tone-mapping, commonly used to ensure a more even brightness distribution. %

\subsection{De-rendering}
Our network architecture is composed of several sub-networks, that predict the different shape, material, and lighting properties of an input image.
From these predictions, we can reconstruct the image with the image formation model described above.
However, as mentioned before, due to the plethora of ambiguities in the rendering function, a simple reconstruction objective alone (\eg $\| \textbf{I} - \hat{\textbf{I}}\|_2^2 $) is not sufficient to learn a meaningful decomposition.
To overcome this challenge, we propose a training scheme, with two additional objectives that regularize the learning problem and prevent degenerate solutions.
As training data, we use an unconstrained set of images with associated coarse geometry estimates.
We use the coarse geometry to generate further coarse estimates for the intrinsic components, which we use in auxiliary loss terms.
These coarse constraints force the model to predict a semantically correct disentanglement on a global level.

\topic{Extracting Coarse Light \& Albedo.}
Because coarse shape (depth map $D_c$ and its normal map $N_c$) alone does not suffice to constrain the decomposition, we also compute coarse light and albedo estimates from the geometry information through two optimization steps.

As we only need coarse estimates of the intrinsic components, we can make the simplifying assumption that the per-pixel coarse brightness $B \in [0, 1]^{H \times W}$ (computed in HSV color space) of the input image is proportional to the combination of ambient and diffuse shading, and discard specular lighting effects.
This translates to an albedo map with constant brightness.
Given light information, we can obtain the relevant shading map from the coarse geometry.
Therefore, we optimize the coarse light parameters $L_c = (s_{\mathrm{amb}, c}, s_{\mathrm{dir}, c}, l_c)$ such that the aggregated shading map corresponds to the brightness of the input image.
\begin{equation}\label{eq:opt_normals}
    \argmin_{L_c} \sum_{u\in \Omega} \left( 2B_u - 
    (s_{\mathrm{amb}, c} + s_{\mathrm{dir}, c} N_{c,u}^T  l_c) \right)^2
\end{equation}
We fix the albedo brightness to $\frac{1}{2}$ to avoid color saturation effects and consequently add a scaling factor of $2$ for $B$.
Here, $N_c$ is the coarse normal map.
With this light estimate, an initial albedo estimate $\tilde{A}_c$ can be obtained by inverting the shading equation:
\begin{equation}\label{eq:coarse_albedo}
    \tilde{A}_{c,u} = \mathbf{I}_u  \left(s_{\mathrm{amb}, c} + s_{\mathrm{dir}, c} N_{c,u}^T l_c \right)^{-1}
\end{equation}
However, because of the coarseness of the geometry and no modeling of specularity effects, an estimate using this formulation alone will contain many artifacts.
To regularize the estimate $\tilde{A}_c$ we refine it using another optimization step.
Similar to the constraints used in the intrinsic image decomposition literature~\cite{xu2012structure,shen2013intrinsic}, we apply total variation regularization (TV) on the albedo as well as a data term that retains the image gradients (\ie edges):
\begin{equation}\label{eq:opt_albedo}
\begin{split}
    \argmin_{A_c} \|\delta_xA_c - \delta_x\tilde{A}_c\|_2 + 
    \|\delta_yA_c - \delta_y\tilde{A}_c\|_2 + \\ 
    \lambda_{\mathrm{TV}} \left(\|\delta_x A_c\|_1 + \|\delta_y A_c\|_1\right) \, .
\end{split}
\end{equation}
We use $\delta_x$ and $\delta_y$ to signify the computation of image gradients, which can, for example, be obtained by applying the Sobel operator to the image.
We obtain $L_c$ and $A_c$ by optimizing \cref{eq:opt_normals} and \cref{eq:opt_albedo} respectively using gradient descent, which takes less than a second and can be precomputed for each image (see \cref{fig:coarse_geometry}).

\topic{Learning to De-render.}
We use three different neural networks to predict the intrinsic components from the input image $\mathbf{I}_\mathrm{in}$.
A shape network $\Phi_{\mathrm{shape}}$ predicts both the depth map $D, D_u \in [d_\mathrm{min}, d_\mathrm{max}]$ and the normal refinement map $N_{\mathrm{ref}}$, which is normalized after prediction and used to obtain the final normal map $N$ with \cref{eq:normal_ref}.
The albedo network $\Phi_{\mathrm{albedo}}$ predicts the albedo map $A, A_u \in [0, 1]$, and the light network $\Phi_{\mathrm{light}}$ predicts the light parameters $s_\mathrm{amb}, s_\mathrm{dir} \in [0, 1]$, as well as, shininess $\alpha  \in [0, \alpha_\mathrm{max}]$ and specular intensity $a_\mathrm{spec} \in [a^\mathrm{spec}_\mathrm{min}, a^\mathrm{spec}_\mathrm{max}]$.

We train our model using complementary losses on the decomposition and on the rendered image.
This makes the network adhere to globally accurate components, while achieving more detailed reconstructions.
The loss is computed using the (precomputed) coarse shape, albedo, and light information as pseudo supervision.
\begin{equation}
\begin{split}
    \mathcal{L}_c = \sum_{u \in \Omega} \lambda_D \|D_u - D_{c,u}\|_1 -
    \lambda_N N_u^T N_{c,u} +\\ \lambda_A \|A_u - A_{c,u}\|_1 + \lambda_L \|L - L_c\|_2^2 
\end{split}
\end{equation}

Additionally, there are two losses on the rendered image.
First, we apply a reconstruction loss between the rendered and the input image to train our model to capture all local details in the decomposition.
Specifically, this loss term is computed from the combination of a per-pixel $\text{L1}$ loss and the patch-based structural similarity score $\text{SSIM}(\textbf{I}, \hat{\textbf{I}})$~\cite{wang2004image}.
\begin{equation}
\begin{split}
    \mathcal{L}_\mathrm{rec} = \frac{1}{|\Omega|} \sum_{u\in\Omega}\|\mathbf{I}_u - \mathbf{\hat{I}}_u\|_1 + 
    \frac{1}{2}\left(1 - \text{SSIM}(\mathbf{I}, \mathbf{\hat{I}})\right)
\end{split}
\end{equation}

While the reconstruction loss gives a very strong training signal, there often remains some ambiguity, in that, given fixed light, certain details can be modeled either by the material (light independent) or the shape component (light dependent).
Such mistakes only become apparent when we render the image under new lighting conditions $L^\prime$ (mainly influenced by the direction $l^\prime$) . 

To ensure that we achieve a semantically correct decomposition, we therefore also introduce an adversarial formulation.
Concretely, we render two images in each forward pass: one with the predicted lighting conditions, denoted as $\mathbf{\hat{I}}$, which is also used in the reconstruction loss term, and one with randomly sampled lighting conditions, denoted as $\mathbf{\hat{I}}^\prime$.
We then train a discriminator network $\Phi_\mathrm{disc} \in \mathbb{R}$ to score, whether an image was rendered using the original lighting conditions or whether it was re-lit. For this we use the discriminator from LSGAN~\cite{mao2017least}.
Using the reconstructed image $\mathbf{\hat{I}}$ instead of the original image $\mathbf{I}_\mathrm{in}$ as positive example when training $\Phi_\mathrm{disc}$, has the advantage that the network cannot use artifacts from the image formation models as hints as to whether the image was re-lit or not, as both, real and fake examples come from the same pipeline.
The loss term on the relit image is computed as:
$ \mathcal{L}_\mathrm{gan} = (1 - \Phi_\mathrm{disc}(\mathbf{\hat{I}}^\prime))^2$. %
We can then train the whole model end-to-end using 
\begin{equation}
    \mathcal{L} = \mathcal{L}_c + \lambda_\mathrm{rec}\mathcal{L}_\mathrm{rec} + \lambda_\mathrm{gan} \mathcal{L}_\mathrm{gan} \, ,
\end{equation}
to learn to de-render an image into its intrinsic components.

\topic{Refinement.}
While simplifying the specularity model to two scalars allows for stable training, it can be limiting when there are large differences in material properties across the object.
To alleviate this issue, similar to the way we allow the normals to deviate from the underlying shape via a refinement map, we predict a  per-pixel specularity refinement map $I_\mathrm{spec\_ref}(\hat{\textbf{I}})$ from the \emph{output} image.
We then multiply $I_\mathrm{spec\_ref}$ with the specularity term and re-compose the image. 
\section{Experiments}

We conduct extensive experiments to evaluate our method and its individual components.
\subsection{Datasets and Metrics}

We use three different datasets to cover a wide variety of objects: faces,  a collection of common objects in the wild, and a new synthetic and photo-realistic test set with ground truth annotations. Please see the supplement for all details.

\topic{CelebA-HQ} \cite{karras2017progressive}
is a large-scale human face dataset, consisting of 30k high-resolution portrait pictures of celebrities.
We roughly crop out the face area and use the corresponding train/val/test split of the CelebA dataset.
To obtain the rough initial geometry estimate $D_c$, we use \cite{Wu_2020_CVPR} at a reduced resolution of $64 \times 64$.

\begin{figure}
  \centering
  \includegraphics[width=\linewidth]{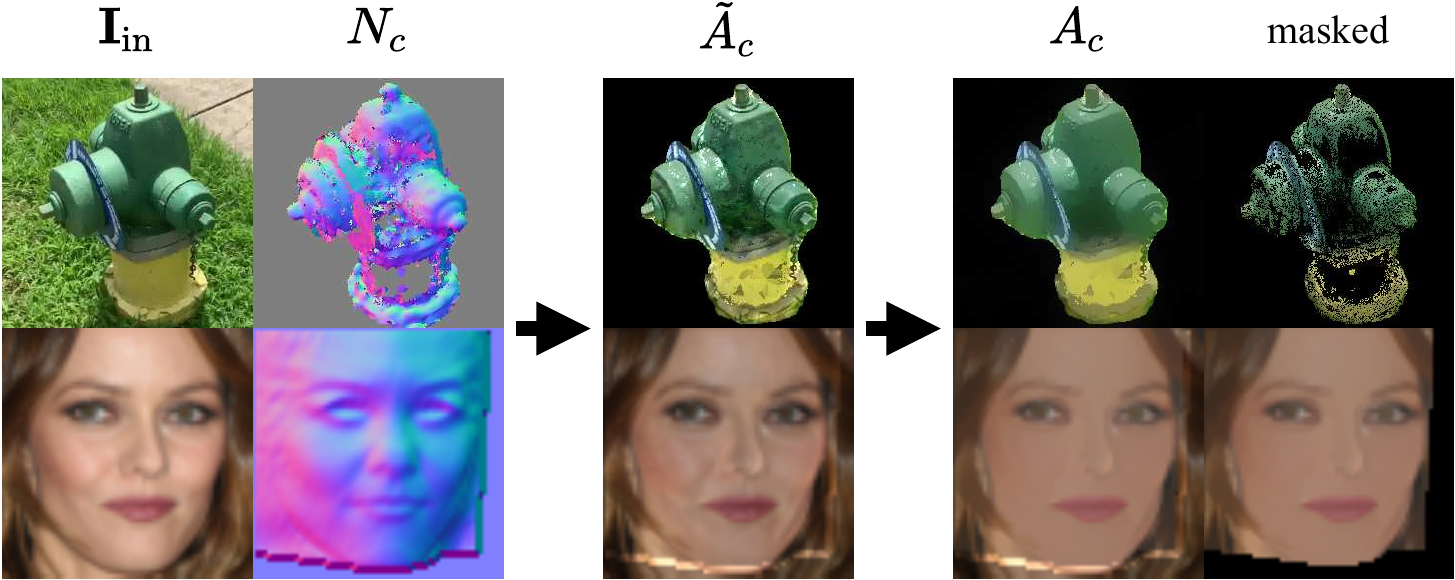}
  \caption{\textbf{Coarse Albedo Optimization.} For training, we precompute coarse albedo estimates $A_c$ from the input image $\mathbf{I}_\mathrm{in}$ and the coarse shape estimate $N_c$. We first approximate the light by assuming a Lambertian shading model and inverting it. Subsequently, we apply a smoothing optimization to remove artifacts.}
   \vspace{-.25cm}

  \label{fig:coarse_geometry}
\end{figure}

\topic{Co3D}  \cite{reizenstein2021common}
is a collection of nearly 19,000 videos capturing objects from 50 MS-COCO~\cite{lin2014microsoft} categories, that come with per-frame depth, camera pose data, and reconstructed sparse point clouds.
First, we use the Point Cloud Library~\cite{Rusu_ICRA2011_PCL} to compute surface normals from the point clouds.
The resulting depth and normal maps are very sparse (see \cref{fig:coarse_geometry}).
We select a subset of the categories and obtain 23895 training and 2817 testing images.

\topic{\dataset}
 (\datasetlong) is a test set we have created to allow for quantitatively evaluation of image decomposition methods.
This is necessary as there does not exist a dataset that combines photorealistic images with precise image decomposition ground truth.
We hand-select 10 freely available and photorealistic 3D scenes for the Blender 3D modeling software\footnote{\url{https://blender.org}} and define 4 different camera views for each.
Additionally to the rendered image, we also save the diffuse albedo map, normal map, and foreground mask.
We do not use this dataset of 40 images only for testing.

\topic{Post-processing for Training.}
For every image, we compute the normal map from the rough initial depth estimate and optimize a light and albedo approximation (see \cref{fig:coarse_geometry}).

We train one model for the \textbf{CelebA-HQ} dataset and one model for the \textbf{Co3D} dataset. 
All hyper-parameters are the same for both models (see supplementary material) except for $\lambda_L=0$ on \textbf{Co3D},
as the geometry is very sparse and the light estimate is often not accurate enough, hindering the convergence of the training.

\topic{Metrics.}
As other methods use different image formation models, and thus obtain a different representation for shading, it is not possible to directly compare shading maps.
This constrains us to quantitative evaluation of normal and albedo map only.
Aside from the common $\mathcal{L}_1$, $\mathcal{L}_2$, and SSIM error metrics, we use mean angle deviation in degrees $\text{DIA}(N, \hat{N}) = \frac{1}{|\Omega|}\sum_{u\in\Omega}\cos^{-1}(N_u^T  \hat{N}_u)$ for normals, and the scale-invariant error $\text{SIE}(A, \hat{A}) = \frac{1}{|\Omega|}\sum_{u\in\Omega}\|A_u - \mu_A - (\hat{A} - \mu_{\hat{A}})\|_2^2$ for albedo, as it can be estimated only up to a constant scale factor.
Here, $\mu_A$ is the average albedo across the whole image $\mu_A = \frac{1}{|\Omega|}\sum_{u\in\Omega}A_u$.

\subsection{Results}

\begin{figure*}
  \centering
  \includegraphics[width=.85\linewidth]{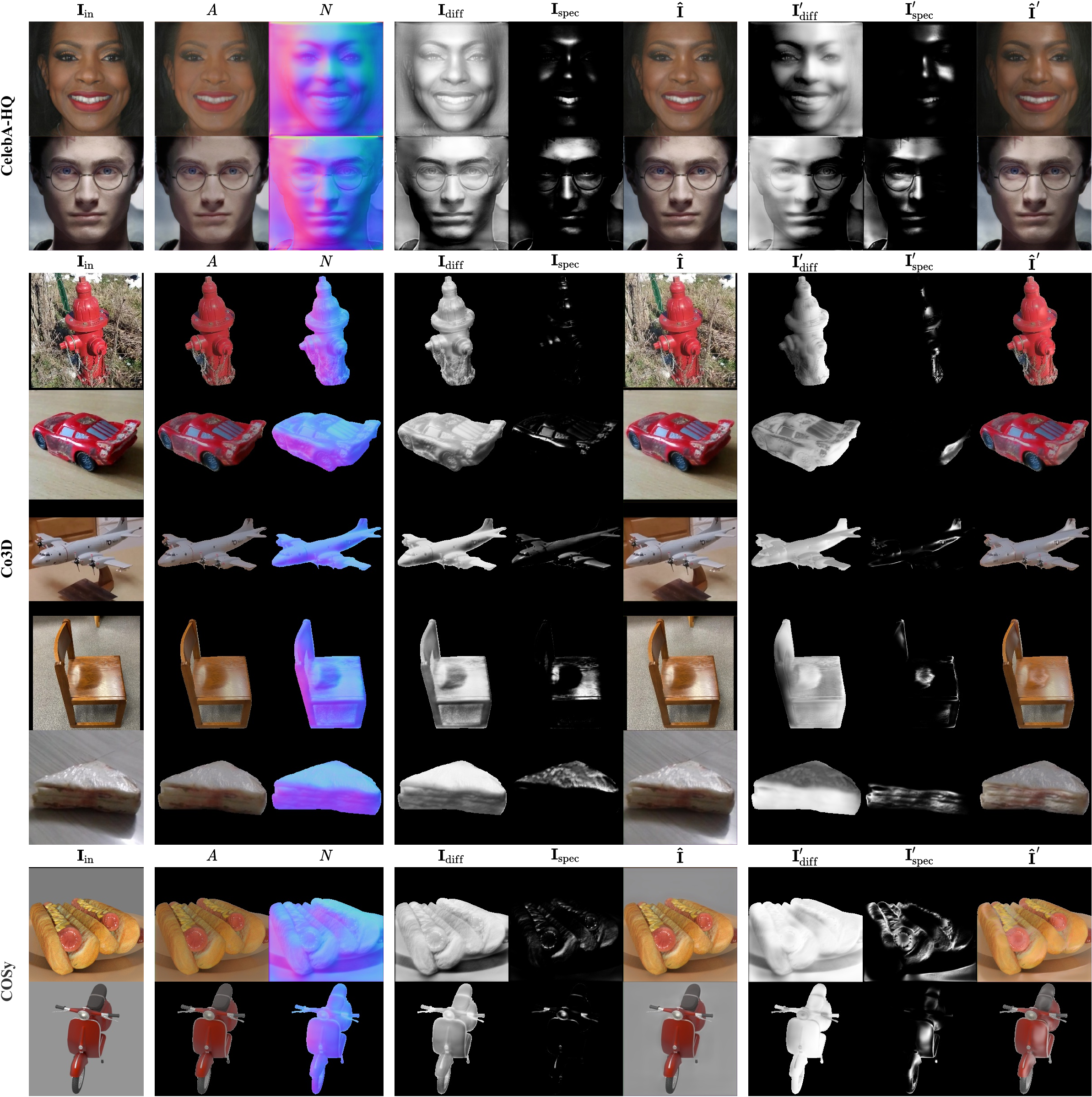}
  \caption{\textbf{Qualitative results.} We train a model on each of the two datasets \textbf{CelebA-HQ} and \textbf{Co3D}, and show the respective decomposition results from the test sets. 
  To highlight the generalization capabilities, we also apply the \textbf{Co3D} model to samples from our synthetic test set \textbf{COSy}. Every row contains the input image $\mathbf{I}_\mathrm{in}$, predicted albedo $A$ and normals $N$, diffuse shading map $\mathbf{I}_\mathrm{diff}$, specular shading map $\mathbf{I}_\mathrm{spec}$ and reconstructed image $\mathbf{\hat{I}}$. Further, we also show the shading maps ($\mathbf{I}_\mathrm{diff}^\prime$, $\mathbf{I}_\mathrm{spec}^\prime$) and reconstructed image $\mathbf{\hat{I}}^\prime$ under new lighting conditions.
  Our model achieves a high level of detail on shape and material reconstruction and convincing relighting results.
  }
   \vspace{-.25cm}
  \label{fig:combined_results}
\end{figure*}

\topic{Qualitative Evaluation.}
To demonstrate the capabilities of our method, we first evaluate it on a diverse selection of samples, as shown in \cref{fig:combined_results}.
Regardless of the category and background, we obtain globally correct results with a very high fidelity.
Critically, even though the \dataset dataset object categories are \emph{not} part of the training categories, we observe the same level of detail.
This demonstrates our method's generalization capabilities to novel objects and categories.
In addition to the decomposition results, we also show that our method produces realistic images and shading maps when changing the light.

\begin{figure}
  \centering
  \includegraphics[width=\linewidth]{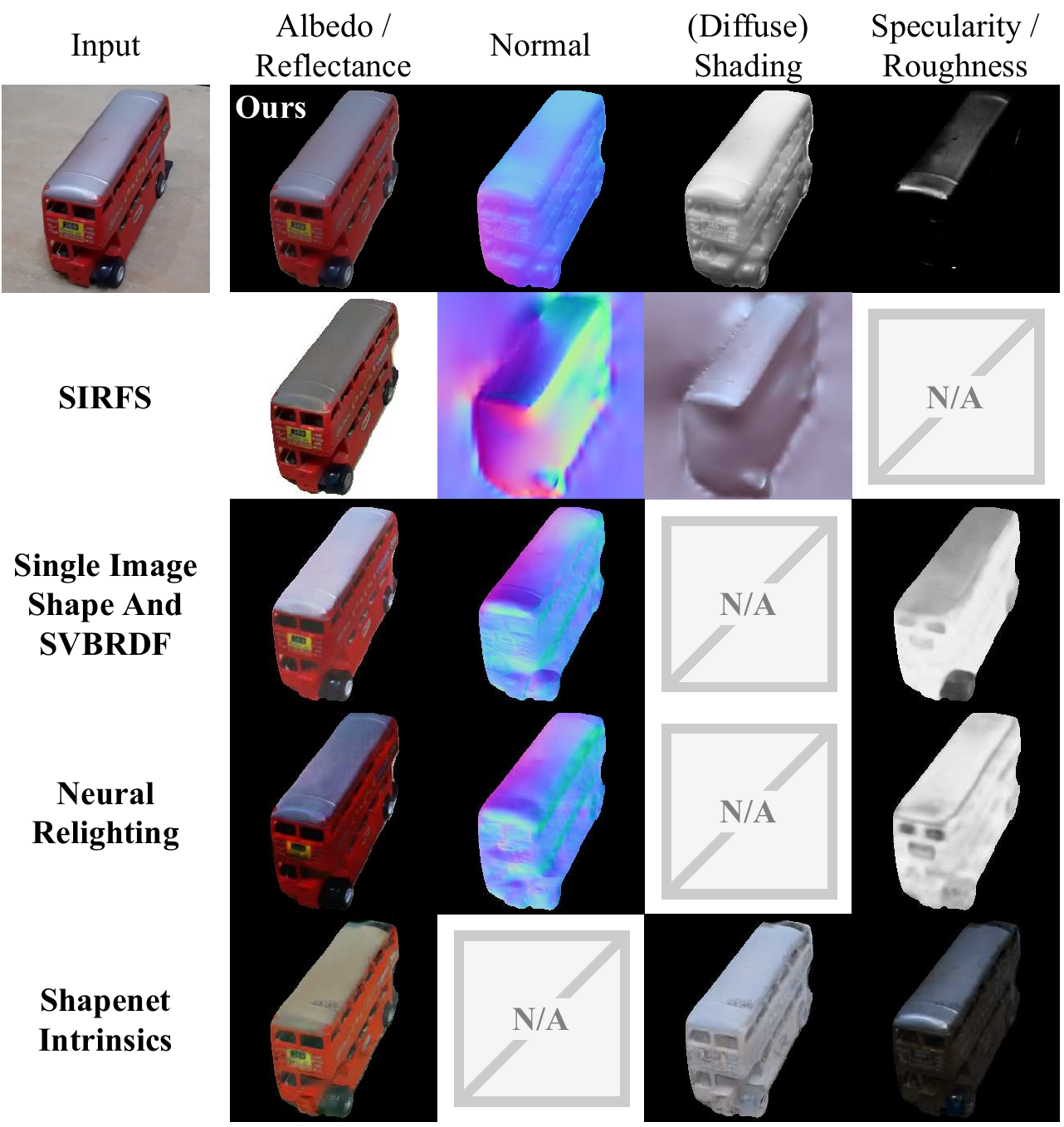}
  \caption{\textbf{Qualitative comparison with state of the art.} We show superior image decomposition results compared to SIRFS \cite{BarronTPAMI2015}, Single Image Shape and SVBRDF \cite{li2018learning}, Neural Relighting \cite{sang2020single}, and ShapeNet Intrinsics \cite{ShapenetIntrinsics}. }
  \label{fig:co3d_comparison}
\end{figure}

Further, we compare our results with state-of-the-art methods for intrinsic image decomposition \cite{BarronTPAMI2015,li2018learning,sang2020single,ShapenetIntrinsics} in \cref{fig:co3d_comparison}.
All methods are able to predict reasonable albedo maps, that capture the major color components.
However, the albedo maps of \cite{BarronTPAMI2015,li2018learning,sang2020single} still contain color gradients and lighting effects around the edges and corners.
\cite{ShapenetIntrinsics} is able to remove almost all specular components of the roof but introduces artifacts, for example, at the top of the roof.
Our method successfully removes shading effects and does not contain artifacts.
For normal prediction, \cite{BarronTPAMI2015} does not capture the shape of the object, nor fine details.
Although \cite{li2018learning} and \cite{sang2020single} predict seemingly detailed normal maps, closer inspection shows that they are not physically grounded (\eg the normals on the windows point upwards).
Our normal map is both detailed and it adheres to the global shape.
Finally, our shading maps are computed from the normal map and additional material properties. 
This is why they are similarly detailed and based on physical properties.
Diffuse and specular effects are captured correctly.
The diffuse shading map predicted by \cite{ShapenetIntrinsics} is very detailed, but does not capture the light direction.

\topic{Quantitative Evaluation.}
As neither \textbf{CelebA-HQ} nor \textbf{Co3D} contain explicit, dense ground truth for the different intrinsic image components, we apply the \textbf{Co3D} model on our newly introduced \textbf{\dataset} dataset to perform quantitative evaluations.
We compare against state-of-the-art image decomposition methods \cite{BarronTPAMI2015, li2018learning, sang2020single,ShapenetIntrinsics}, as shown in \cref{tab:synthetic_dataset}.
Across all metrics, we achieve best accuracy on normal and albedo extraction.
The fact, that our method (like the others) was not trained on this test set, highlights its strong generalization capabilities.

\begin{table}[t]
  \centering
  \footnotesize
  \setlength{\tabcolsep}{5pt}
\begin{tabular}{lccccc}\toprule
& \multicolumn{2}{c}{Normal $N$} & &  \multicolumn{2}{c}{Albedo $A$} \\ \cmidrule{2-3}\cmidrule{5-6}
Model & \textbf{MSE} $\downarrow$ & \textbf{DIA} $\downarrow$ & & \textbf{SIE} $\downarrow$ & \textbf{SSIM} $\uparrow$ \\
\midrule
SIRFS \cite{BarronTPAMI2015}                 & 0.331      & 52.994 & & 0.113      & 0.724     \\
ShapeNet-Intr.\ \cite{ShapenetIntrinsics}    & N/A        & N/A    & & 0.114      & 0.726     \\
SISaSVBRDF\cite{li2018learning}              & 0.288      & 42.801 & & 0.112      & 0.752     \\
Neur.\ Rel.\ \cite{sang2020single}           & 0.228      & 41.603 & & 0.093      & 0.723     \\ \midrule
\textbf{Ours}                                & \textbf{0.173} & \textbf{37.807} & & \textbf{0.075} & \textbf{0.760}    \\ \bottomrule
\end{tabular}
\caption{\textbf{Comparison with the state of the art.} We show good improvements over previous methods on the \dataset dataset. \cite{li2018learning} and \cite{sang2020single} were trained using flash photographs.}
\label{tab:synthetic_dataset}
\end{table}

\begin{table}[t]
  \centering
  \footnotesize
  \setlength{\tabcolsep}{3.5pt}
\begin{tabular}{lcccccccc} \toprule
& \multicolumn{2}{c}{Normal $N$} & &  \multicolumn{2}{c}{Albedo $A$} & &  \multicolumn{2}{c}{Specular $I_\mathrm{spec}$} \\ \cmidrule{2-3}\cmidrule{5-6}\cmidrule{8-9}
Model & \textbf{MSE} $\downarrow$ & \textbf{DIA} $\downarrow$ & & \textbf{SIE} $\downarrow$ & \textbf{SSIM} $\uparrow$ & & \textbf{MSE} $\downarrow$ & \textbf{L1} $\downarrow$\\
\midrule
No Albedo & 0.162 & 36.5 & & 0.088 & 0.750 & & 0.124 & 0.077 \\
No Shape  & 0.506 & 68.7 & & 0.079 & 0.757 & & 0.108 & 0.058 \\
No GAN    & 0.169 & 37.2 & & 0.075 & 0.762 & & 0.123 & 0.073 \\
\midrule
Ours      & 0.173 & 37.8 & & 0.075 & 0.760 & & 0.112 & 0.059 \\\bottomrule
\end{tabular}
\caption{\textbf{Ablation.} Results on the \dataset dataset when deactivating components of our model. Concretely, we set the $\lambda=0$ coefficient for the respective loss term and then do a full training run.}
\vspace{-.25cm}
\label{tab:ablation_synthetic}
\end{table}   %

\begin{figure}
  \centering
  \includegraphics[width=\linewidth]{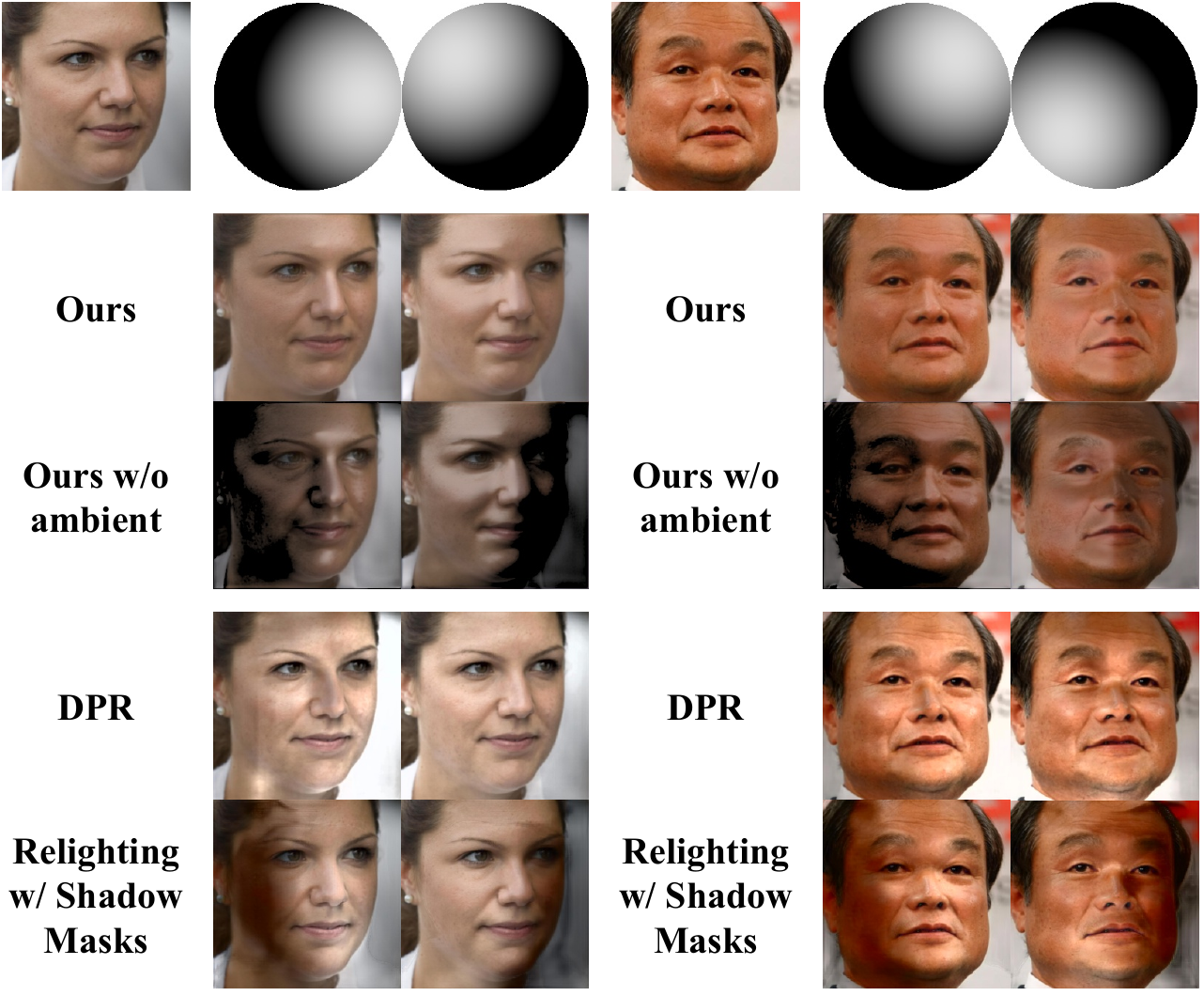}
  \caption{\textbf{Qualitative comparison with face relighting methods}. We use our model to relight images from the \textbf{CelebA-HQ} test set and compare with state-of-the-art face relighting methods. Our method shows better color accuracy and more robustness.}
   \vspace{-.25cm}
  \label{fig:face_relighting}
\end{figure}

\topic{Single Image Relighting.}
To demonstrate the usefulness of de-rendering, we perform relighting on the \textbf{CelebA-HQ} dataset. \cref{fig:face_relighting} shows comparison of our method with state-of-the-art face relighting methods \cite{zhou2019deep} and \cite{hou2021towards}.
As a result of the underlying explicit image formation model, our method produces visually correct relighting results, 
which are more color accurate than \cite{zhou2019deep} and contain much fewer artifacts than \cite{hou2021towards}.
This demonstrates that our method can not only perform very well in intrinsic image decomposition, but also compete with methods, that were specifically designed for certain sub-tasks.

\subsection{Ablation Study and Analysis}
We also conduct several ablation studies and further analyses of the impact of the individual model components. 

\topic{Loss Components.}
At the heart of our method is the combination of three losses: a coarse loss on the different intrinsic components, a reconstruction loss, and a discriminator loss.
We deactivate each component and then evaluate the resulting models on \dataset, as shown in \cref{tab:ablation_synthetic}.

When deactivating the albedo and shape losses ($\lambda_A=0$ and $\lambda_D=\lambda_N=0$ respectively), the predictions of the respective components become significantly worse.
The discriminator loss does not have a large influence on the quality of the albedo and normal accuracy, however, it stabilizes the accuracy of the specular shading map.

\topic{Geometry \& Albedo Improvement.}
\cref{fig:coarse_geometry} compares on two (test-set) examples the coarse input during training and the prediction of the trained model.
This is to verify that albedo and normal map predictions achieve a significantly higher level of detail and completeness compared to their initial, coarse counterparts that are used to supervise the training process.
This is a result of the reconstruction and GAN losses and the explicit image formation model.

\begin{figure}
  \centering
  \includegraphics[width=\linewidth]{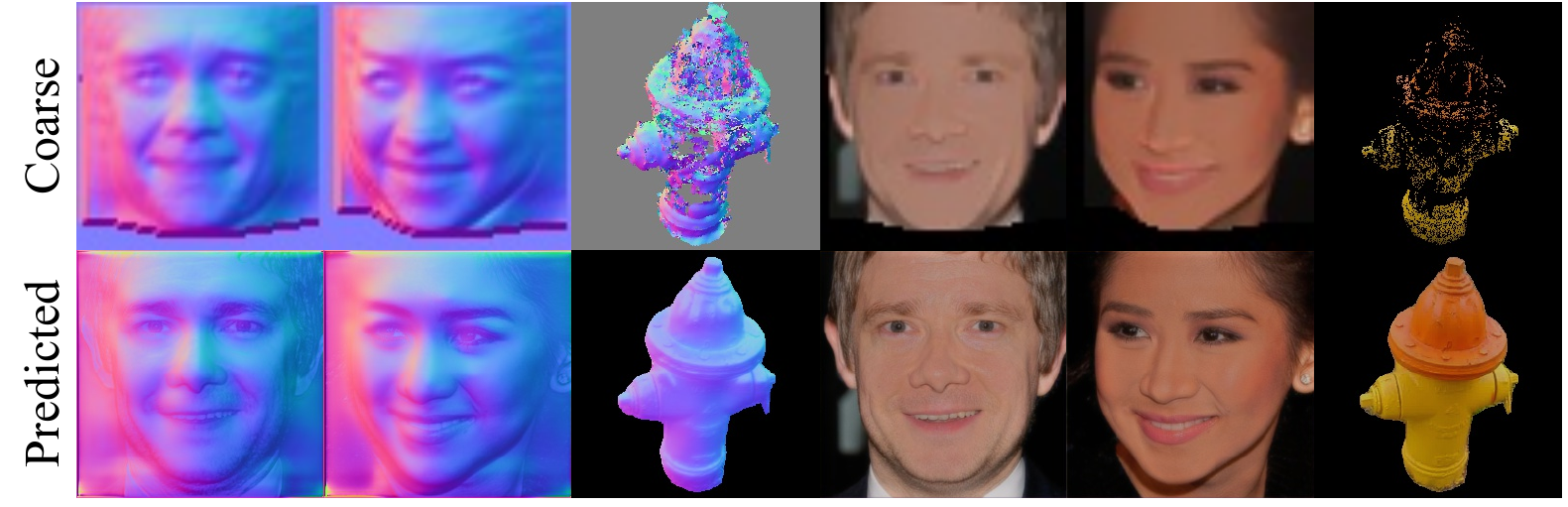}
  \caption{\textbf{Coarse estimates vs.\ prediction.} We compare the rough shape and albedo estimates that we use during training with predictions from the final model and show large improvements. Note that the rough estimates are not available during test time.}
  \vspace{-.25cm}
  \label{fig:ablation_nr}
\end{figure}

\begin{figure}
  \centering
  \includegraphics[width=\linewidth]{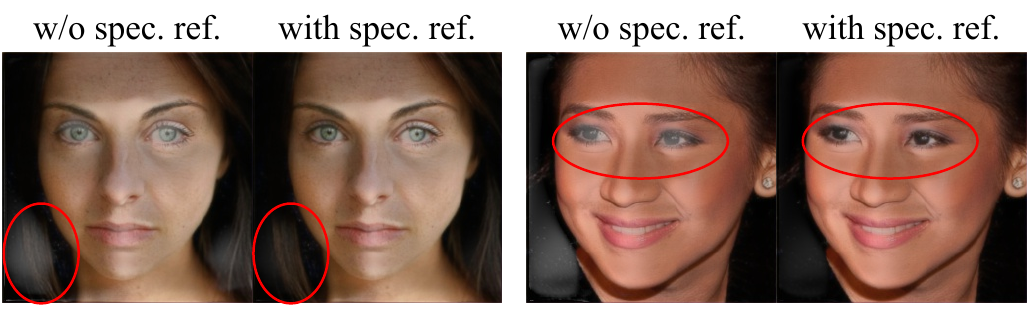}
  \caption{\textbf{Specular Refinement.} The specular refinement network improves regions where the complexity of underlying material is difficult to describe explicitly with the lighting model.}
  \vspace{-.25cm}
  \label{fig:ablation_spec_ref}
\end{figure}

\topic{Specular Refinement.}
\cref{fig:ablation_spec_ref} demonstrates the specular refinement on two re-lit portrait images.
The assumption of shared specular parameters for the entire image, can sometimes lead to specular artifacts in complex regions, which is especially important during relighting.
The network effectively removes artifacts both on the hair and around the eyes, leading to a more realistic output.

\section{Conclusions}

We have presented a method that can factorize in-the-wild images of objects into their intrinsic components shape, material, and lighting.
Our proposed learning pipeline does not rely synthetic datasets and only uses sparse geometry estimates during training, which can be obtained using off-the-shelf unsupervised methods.
Through a series of ablation studies, we have demonstrated the importance of the different components of our method, particularly the coarse losses.
The proposed method achieves high accuracy for all intrinsic components, both on in- and out-of-distribution images, which we measure on our newly introduced synthetic image decomposition test set that we hope will become a new benchmark for de-rendering images in the wild.

\footnotesize{
\paragraph{Acknowledgments}
Shangzhe Wu is supported by Meta Research.
Christian Rupprecht is supported by Innovate UK (project 71653) on behalf of UK Research and Innovation (UKRI) and the Department of Engineering Science at the University of Oxford.
}

{\small
\bibliographystyle{ieee_fullname}
\bibliography{ms}
}

\clearpage
\section{Supplementary Material}

\subsection{Additional Results}
\begin{figure*}
  \centering
  \includegraphics[width=\linewidth]{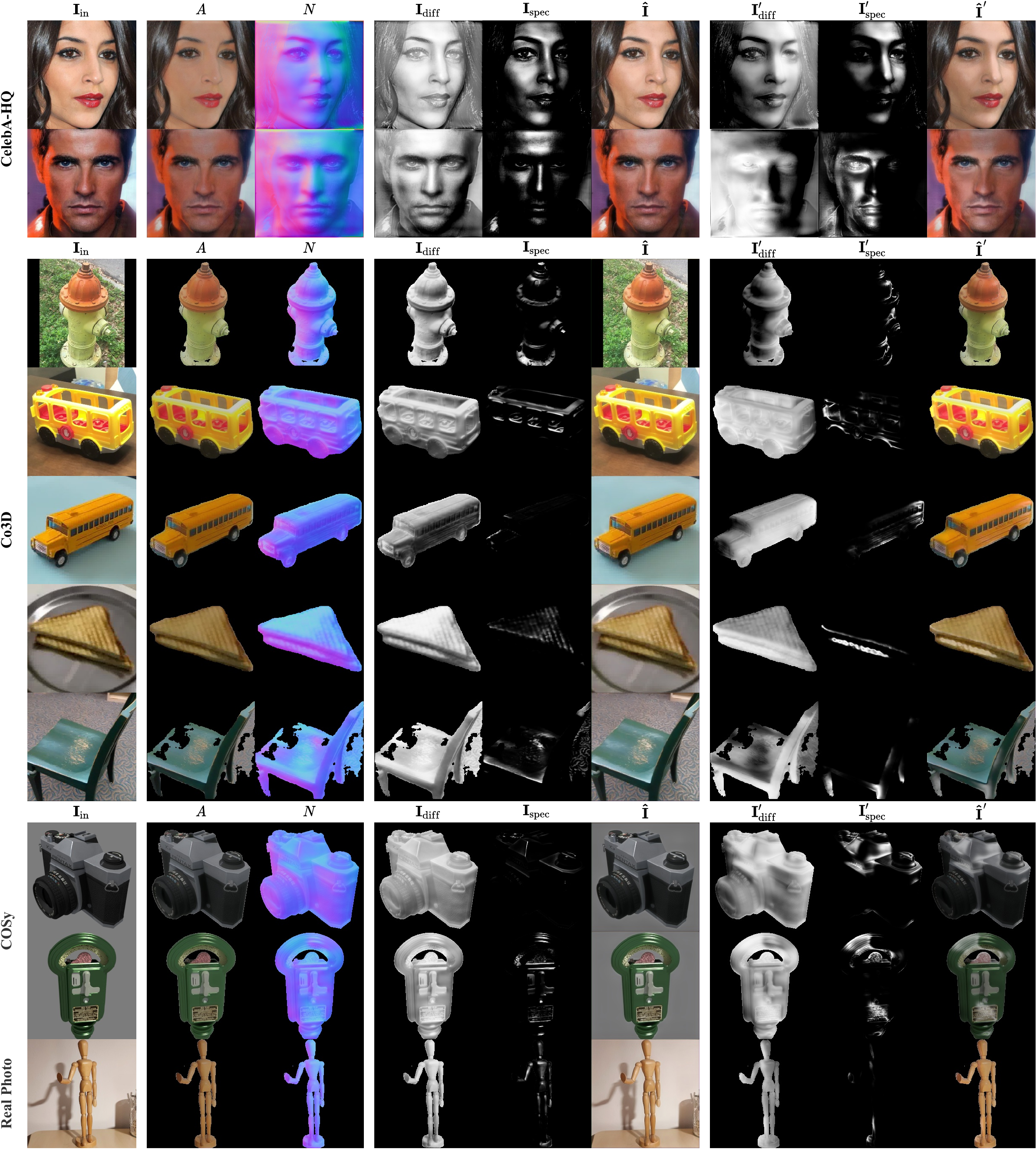}
  \caption{\textbf{Additional qualitative results.} Corresponding to Fig. 4 in the main paper.
  }
  \label{fig:add_results}
\end{figure*}

\begin{figure*}
  \centering
  \includegraphics[width=\linewidth]{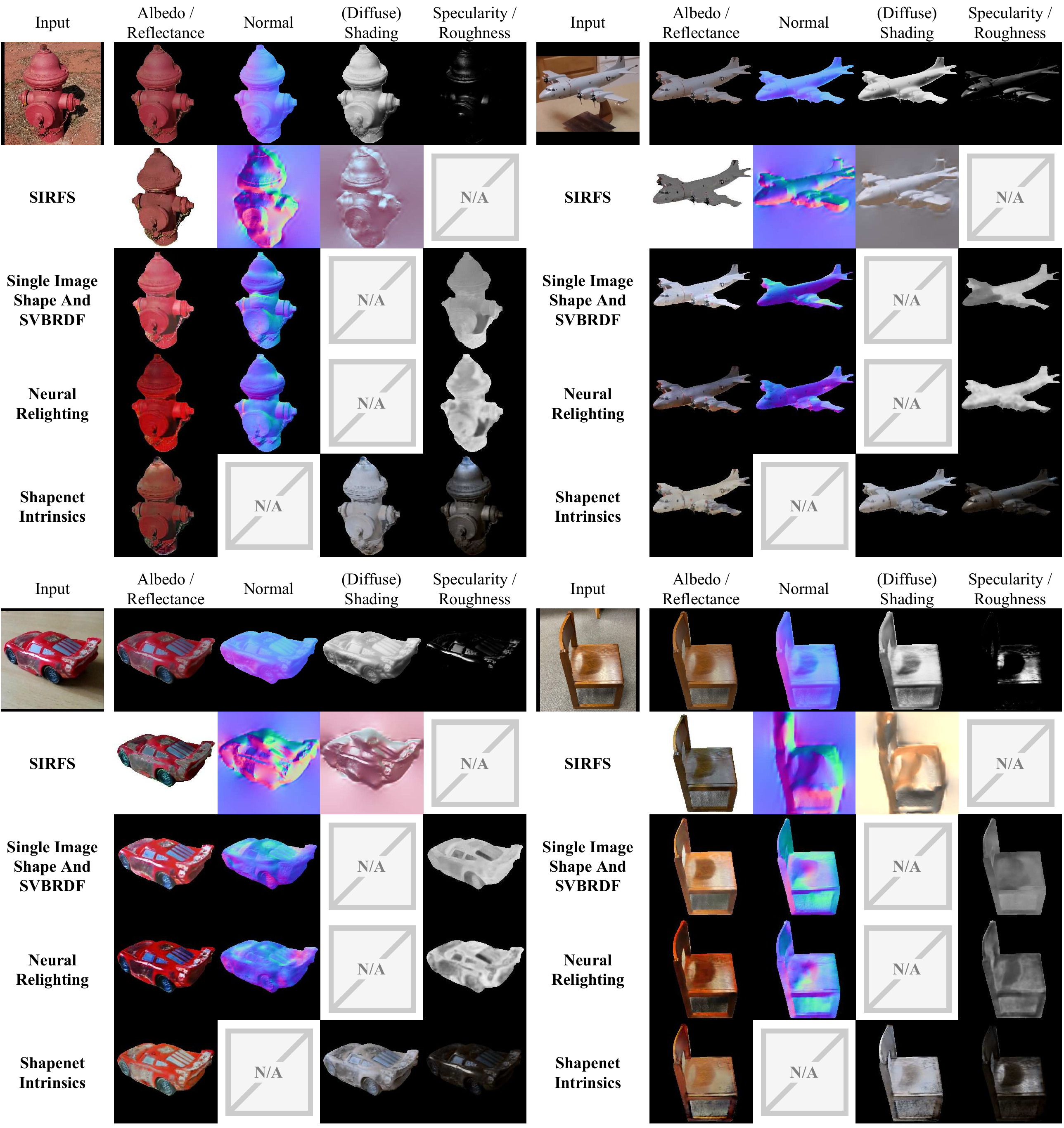}
  \caption{\textbf{Additional qualitative comparison with state of the art.} Corresponding to Fig. 6 in the main paper.}
  \label{fig:add_co3d_comparisons}
\end{figure*}

We show some additional qualitaive results. \Cref{fig:add_results} shows a large number of additional decomposition results.
\Cref{fig:add_co3d_comparisons} shows a large number of additional visual comparisons with other state-of-the-art methods.
\cref{tab:reb_bias} analyses the effects of bias in the training data, which might get introduced as a result of our coarse shape, light, and material estimation.

\begin{table}[b]
  \centering
  \footnotesize
  \setlength{\tabcolsep}{3.0pt}
\begin{tabular}{lcccccccc} \toprule
& \multicolumn{2}{c}{Normal $N$} & &  \multicolumn{2}{c}{Albedo $A$} & &  \multicolumn{2}{c}{Specular $I_\mathrm{spec}$} \\ \cmidrule{2-3}\cmidrule{5-6}\cmidrule{8-9}
Add. bias & \textbf{MSE} $\downarrow$ & \textbf{DIA} $\downarrow$ & & \textbf{SIE} $\downarrow$ & \textbf{SSIM} $\uparrow$ & & \textbf{MSE} $\downarrow$ & \textbf{L1} $\downarrow$\\
\midrule
None     & 0.173 & 37.8 & & 0.075 & 0.760 & & 0.112 & 0.059 \\
\midrule
Normal $+ 15^\circ$ &  0.184 & 39.4 && 0.077 & 0.759 && 0.124 & 0.077 \\
Brightn. $-0.1$ & 0.171 & 37.7 && 0.075 & 0.766 && 0.115 & 0.059 \\
Brightn. $+0.1$ & 0.183 & 39.1 && 0.077 & 0.716 && 0.122 & 0.065 \\
\bottomrule
\end{tabular}
\caption{\textbf{Effects of bias in training data} Artificial bias introduced for normal angles and the brightness of the coarse albedo estimates.}

\label{tab:reb_bias}
\vspace{-.5cm}
\end{table}

\subsection{Limitations}
While our proposed method achieves good results on a wide variety of data, it also comes with some limitations.

First, when there is no coarse light supervision, the training can become unstable and might, after a long training, suddenly diverge to a state where it ignores specular effects.
This usually shows in $\alpha$ taking maximum value and the $a_\mathrm{spec}$ taking minimum value.

\begin{figure}
  \centering
  \includegraphics[width=\linewidth]{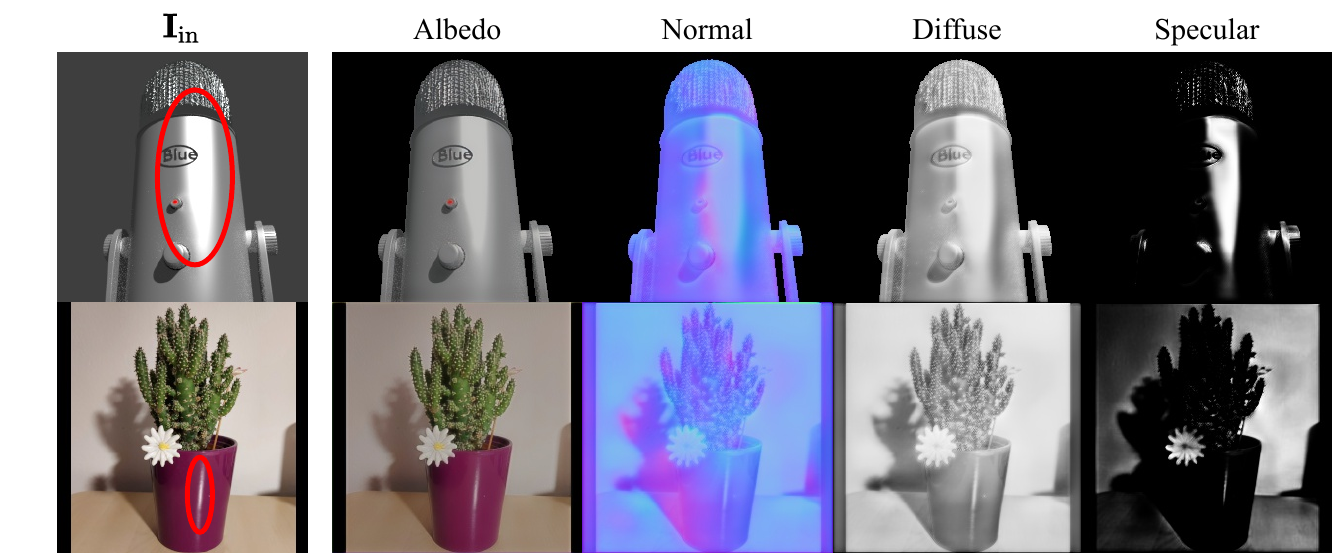}
  \caption{\textbf{Limitations.} Extreme lighting effects with out-of-distribution examples. Strong specular reflections can sometimes bleed into the albedo map. Strong shadows can have an effect on the normal map. 
  }
  \label{fig:limitations}
\end{figure}

Secondly, even though our image formation model is already fairly complex, it can sometimes still not capture all possible lighting effects that can appear in in-the-wild images.
In extreme cases, the model tends  to bake those shading effects into either the albedo or the shading map, as shown for example in \cref{fig:limitations}. Additionally, hard shadows can sometimes have an influence on the normal map, even though the shape underneath is of course unaffected by the shadow of another object falling onto it (see second row of \cref{fig:limitations}). We expect that an explicit modeling of shadows could alleviate this problem.

\subsection{Broader Impact \& Ethics}
The goal of our method is to learn a model that lifts images of objects from 2D into a disentangled 3D representation. 
We expect this method to be useful for the increasing amount of XR and VFX application in consumer devices. 
Additionally, learning to de-render images is in itself a challenging computer vision task as the model needs to discover robust and general visual representations. 

Alongside the paper, we are releasing a synthetic dataset of 3D objects for future benchmarking under an open, non-commerical license. We expect this to contribute to the field as it establishes a test set with ground truth annotations that was previously not available to researchers.

As the main focus of this work concerns general objects, potential negative societal implications are low. However, we include an evaluation on CelebA-HQ \cite{karras2017progressive}, which is a dataset of images of faces of celebrities. The copyright situation is unclear as the dataset was scraped from the internet and naturally contains person identifying information (faces). The dataset contains biases in many forms (celebrities are not representative of the general population, most images are professional photographs, general focus on celebrities from western countries etc.). Models trained on this data can reflect these biases and thus should only be used for research purposes. We include these results to show a comparison to methods that are specialized for human faces while our method can be trained on an object-centric dataset. 

As described in the previous section, our method still has some limitations and thus should not be used in critical applications.

\subsection{Technical Details}

\topic{3D Geometry.}
The coordinate system used for representing the normal maps has its x-axis pointing left, y-axis pointing upwards, and z-axis pointing towards the viewer.
We visualize normal maps by first scaling and translating the value range from $[-1, -1]$ to $[0, 1]$ and then directly converting the xyz coordinates to rgb colors.

The lighting direction is modeled by a vector pointing from the model to the light source. 
Note, that the lighting is modeled to be relative to the camera position
Therefore, the viewing direction, relevant for specular effects, is always constant at $v = (0, 0, 1)^T$.

To obtain the normal map $N_\mathrm{D}$ from the depth map $D$, which is used in Eq.\ 1, we compute the 3D tangent plane of a point using the 4-neighborhood of this point in image space. The normal of the tangent plant is then used as the normal at this point.

The coarse normal map $N_c$ is the a key component in the optimization to obtain the coarse albedo and light estimate. 
It is usually either given at a lower resolution than the input image, or it is sparse.
In the case of lower resolution, we downsample the input image to match the resolution of the coarse normal map.
When computing the coarse losses, we upsample the coarse normal and albedo map again.

In the case that $N_c$ is sparse, we fill it in using nearest neighbor interpolation before computing the initial albedo estimate $\tilde{A}_c$.
However, during albedo and light optimization, as well as during the computation of the coarse losses, we only consider valid pixels.

\topic{Light Prediction and Sampling.}
When predicting the light direction, we fix the $z = 1$ (facing from the camera side to the object) and only predict the $x, y \in [-1, 1]$ components.
Afterwards, we normalize the vector.
The reasoning behind this is that there only exist meaningful lighting effects, when the light is not coming from behind the object.
Further, we apply a custom scaling function $f(x) = (\frac{x + 1}{2} (\alpha_\mathrm{max} - 1) + 1)^2$ to obtain $\alpha$.
We also limit the range of $a_\mathrm{spec}$ to $[a_\mathrm{spec, min}, a_\mathrm{spec, max}]$

The adversarial loss requires images that are relit under random lighting conditions.
To also cover unusual light directions $l^\prime = (x^\prime, y^\prime, z^\prime)$, we again fix $z^\prime = 1$, sample $x^\prime, y^\prime \sim \mathcal{N}(0, \sigma_l)$ and then normalize $l^\prime$.
As strength of ambient and directional lighting only really influences the overall brightness of the image, which is a very easy cue to pick up on for the discriminator, they have to remain within the distribution of the training data.
Therefore, we compute the mean $\mu_\mathrm{amb}$ of $s_\mathrm{amb}$ over the training batch and then sample $s_\mathrm{amb}^\prime \sim \mathcal{N}(\mu_\mathrm{amb}, 0.1)$ (same for $s_\mathrm{dir}^\prime$).

\topic{Implementation.}
All models and data processing steps (except for the Point Cloud Library) are implemented in PyTorch \cite{paszke2019pytorch}.
The Image-to-image networks (shape, albedo, specular refinement) are implemented as auto-encoders with skip-connections, inspired by the U-Net \cite{ronneberger2015u}.
The light network follows a classical encoder architecture.
All of them use the $\text{tanh}$ activation function and the respective outputs gets scaled to the corresponding value range (\eg to $[0, 1]$ for colors).
\cref{tab:arch_e} and \cref{tab:arch_ae} give an overview over the different network configurations.

\topic{Training.}
For all training runs, we use a batch size of 12 on a GPU with 24GB VRAM.
We train the \textbf{CelebA-HQ} model for 30 epochs (approx.\ 60k iterations) and the \textbf{Co3D} model for 10 epochs (approx.\ 20k iterations).
We only use specular refinement for the CelebA-HQ model and train it in a second training stage for $5$ more epochs.
Here, we freeze all other network weights and only use the adversarial loss with a weight of $\lambda_\mathrm{gan}=0.1$.
\cref{tab:hyper_config} shows the exact hyperparameters used for training. On Co3D, we use the objects masks obtained from SfM to mask the prediction during training and inference.

\begin{table}[]
  \centering
  \footnotesize
  \setlength{\tabcolsep}{5pt}
\begin{tabular}{lll}
\toprule
\multicolumn{3}{c}{\textbf{Model Configuration}}\\
\cmidrule{1-3}
Param. & CelebA-HQ & CO3D \\
\midrule
$D$ & $[0.9, 1.1]$ & $[0.9, 1.1]$ \\
$a_\mathrm{spec}$ & $[0.0,  0.5]$ & $[0.1,  0.5]$ \\
$\alpha_\mathrm{max}$ & 64 & 64\\
\bottomrule
\end{tabular}

\begin{tabular}{llll}
\toprule
\multicolumn{4}{c}{\textbf{Training Configuration}}\\
\cmidrule{1-4}
Param. & Value & Param. & Value \\
\midrule
$n$ & 12 & $\lambda_A$ & $1$ \\
$\eta$ & $1e^{-4}$ & $\lambda_L$ & $1${\tiny (CelebA-HQ)} / $0${\tiny (Co3D)}  \\
$\lambda_D$ & $0.5$ & $\lambda_\mathrm{rec}$ & $0.5$ \\
$\lambda_N$ & $1$ & $\lambda_\mathrm{gan}$ & $0.01$ \\
\bottomrule
\end{tabular}

\begin{tabular}{llll}
\toprule
\multicolumn{4}{c}{\textbf{Optimization Configuration}}\\
\cmidrule{1-4}
\multicolumn{2}{c}{\textbf{CelebA-HQ}} & \multicolumn{2}{c}{\textbf{CO3D}}\\
\cmidrule{1-4}
Param.  & Value & Param.  & Value \\
\midrule
$i$ & 100 & $i$ & 100 \\
$\eta_\mathrm{light}$ & 0.01 & $\eta_\mathrm{light}$ & 0.01 \\
$\eta_\mathrm{albedo}$ & $0.04$ & $\eta_\mathrm{albedo}$ & $0.01$ \\
$\lambda_\mathrm{TV}$ & $5$ & $\lambda_\mathrm{TV}$ & $20$\\
\bottomrule
\end{tabular}
\caption{\textbf{Model Configuration and Hyperparameters.} $n$ denotes batch size. $\eta$ denotes the learning rate. $i$ denotes the number of iterations for optimization.}
\label{tab:hyper_config}
\end{table}

\begin{table}[t]
\footnotesize
\begin{center}
\caption{\textbf{Encoder Architecture.} Architecture of the shape $f_S$ and pose network $f_P$. The network follows a convolutional encoder structure. $n$ is the number of parameters predicted by each network.}
\begin{tabular}{lc}
\toprule
 Encoder & Output size \\ \midrule
 Conv(3, 64, 4, 2) + ReLU() & 128 $\times$ 128\\
 Conv(64, 128, 4, 2) + ReLU() & 64 $\times$ 64\\
 Conv(128, 256, 4, 2) + ReLU() & 32 $\times$ 32\\
 Conv(256, 512, 4, 2) + ReLU() & 16 $\times$ 16\\
 Conv(512, 512, 4, 2) + ReLU() & 8 $\times$ 8\\
 Conv(512, 512, 4, 2) + ReLU() & 4 $\times$ 4\\
 Conv(512, 512, 4, 2) + ReLU() & 2 $\times$ 2\\
 Conv(512, 512, 4, 1) + ReLU & 1 $\times$ 1\\ 
 Conv(512, $c_\mathrm{out}$, 1, 1) $\rightarrow$ output & 1 $\times$ 1\\ 
\bottomrule
\end{tabular}
\end{center}
\label{tab:arch_e}
\end{table}

\begin{table*}[t]
\footnotesize
\begin{center}
\caption{\textbf{Auto-Encoder Architecture.} Architecture of $\Phi_\mathrm{shape}$, $\Phi_\mathrm{albedo}$, and $\Phi_\mathrm{spec\_ref}$. The network follows a U-Net structure \cite{ronneberger2015u}. All convolution operators zero-pad the input such that the output has the same resolution.}
\begin{tabular}{lc}
\toprule
 Encoder & Output size \\ \midrule
 Conv(3, 64, 3, 1) + LReLU(0.1) + Conv(64, 64, 3, 1) + LReLU(0.1) + MaxPool(2) & 128 $\times$ 128\\
 Conv(64, 128, 3, 1) + LReLU(0.1) + Conv(128, 128, 3, 1) + LReLU(0.1) + MaxPool(2) & 64 $\times$ 64\\
 Conv(128, 256, 3, 1) + LReLU(0.1) + Conv(256, 256, 3, 1) + LReLU(0.1) + MaxPool(2) & 32 $\times$ 32\\
 Conv(256, 512, 3, 1) + LReLU(0.1) + Conv(512, 512, 3, 1) + LReLU(0.1) + MaxPool(2) & 16 $\times$ 16\\
 Conv(512, 1024, 3, 1) + LReLU(0.1) + Conv(1024, 1024, 3, 1) + LReLU(0.1) + MaxPool(2) & 8 $\times$ 8\\
 Conv(1024, 1024, 3, 1) + LReLU(0.1) + Conv(1024, 1024, 3, 1) + LReLU(0.1) + MaxPool(2) & 4 $\times$ 4\\
 Conv(1024, 1024, 3, 1) + LReLU(0.1) + Conv(1024, 1024, 3, 1) + LReLU(0.1) + MaxPool(2) & 2 $\times$ 2\\
 Conv(1024, 1024, 3, 1) + LReLU(0.1) + Conv(1024, 1024, 3, 1) + LReLU(0.1) & 2 $\times$ 2\\ \midrule \midrule
 Decoder & Output size \\ \midrule
 Conv(1024, 1024, 3, 1) + LReLU(0.1) + Conv(1024, 1024, 3, 1) + LReLU(0.1) + Upsample(2) & 4 $\times$ 4\\
 Conv(1024, 1024, 3, 1) + LReLU(0.1) + Conv(1024, 1024, 3, 1) + LReLU(0.1) + Upsample(2) & 8 $\times$ 8\\
 Conv(1024, 1024, 3, 1) + LReLU(0.1) + Conv(1024, 1024, 3, 1) + LReLU(0.1) + Upsample(2) & 16 $\times$ 16\\
 Conv(1024, 512, 3, 1) + LReLU(0.1) + Conv(512, 512, 3, 1) + LReLU(0.1) + Upsample(2) & 32 $\times$ 32\\
 Conv(512, 256, 3, 1) + LReLU(0.1) + Conv(256, 256, 3, 1) + LReLU(0.1) + Upsample(2) & 64 $\times$ 64\\
 Conv(256, 128, 3, 1) + LReLU(0.1) + Conv(128, 128, 3, 1) + LReLU(0.1) + Upsample(2) & 128 $\times$ 128\\
 Conv(128, 64, 3, 1) + LReLU(0.1) + Conv(64, 64, 3, 1) + LReLU(0.1) + Upsample(2) & 256 $\times$ 256\\
 Conv(64, $c_\mathrm{out}$, 3, 1) + LReLU(0.1) + Conv($c_\mathrm{out}$, $c_\mathrm{out}$, 3, 1) & 256 $\times$ 256\\
\bottomrule
\end{tabular}
\end{center}

\label{tab:arch_ae}
\end{table*}

\topic{COSy Dataset.}
The \textbf{COSy} dataset is built from ten publicly available 3D scenes for the Blender 3D modeling software\footnote{\href{https://www.blender.org/}{https://www.blender.org/}}.
The 3D models can be downloaded from the following links and are available under variants of the Creative Commons license.
Hot Dog\footnote{\href{https://blendswap.com/blend/23962}{https://blendswap.com/blend/23962}}, Accordion\footnote{\href{https://blendswap.com/blend/17099}{https://blendswap.com/blend/17099}}, Wall-Phone\footnote{\href{https://blendswap.com/blend/19579}{https://blendswap.com/blend/19579}}, Hydrant\footnote{\href{https://blendswap.com/blend/8443}{https://blendswap.com/blend/8443}}, Wingback Chair\footnote{\href{https://blendswap.com/blend/12555}{https://blendswap.com/blend/12555}}, Camera\footnote{\href{https://blendswap.com/blend/15833}{https://blendswap.com/blend/15833}}, Toaster\footnote{\href{https://blendswap.com/blend/6231}{https://blendswap.com/blend/6231}}, Scooter\footnote{\href{https://blendswap.com/blend/5256}{https://blendswap.com/blend/5256}}, Microphone\footnote{\href{https://blendswap.com/blend/4145}{https://blendswap.com/blend/4145}}, Parkingmeter\footnote{\href{https://blendswap.com/blend/7714}{https://blendswap.com/blend/7714}}.
We adapt the models to fit our requirements, for example, define camera views and modify the material so that we can extract the diffuse albedo.

\end{document}